\documentclass[11pt]{article}

\usepackage[final]{acl}

\usepackage{times}
\usepackage{latexsym}
\usepackage[T1]{fontenc}
\usepackage[utf8]{inputenc}
\usepackage{microtype}
\usepackage{inconsolata}
\usepackage{graphicx}
\usepackage{booktabs}
\usepackage{multirow}
\usepackage{xcolor}
\usepackage{amsmath}
\usepackage{amssymb}
\usepackage{enumitem}
\usepackage{adjustbox}
\usepackage{pgfplots}
\pgfplotsset{compat=1.18}
\usepgfplotslibrary{groupplots}

\setlist[description]{leftmargin=1.4em, labelindent=0pt, itemsep=1pt, topsep=2pt}
\setlist[enumerate]{leftmargin=1.4em, itemsep=1pt, topsep=2pt}

\title{Natural-Language Temporal Grounding in Hour-Long Videos\\
is a Search Problem: A Benchmark and Empirical Decomposition}

\author{Sukmin Seo$^\ast$\\
  NAVER Cloud AI\\
  \texttt{mellonggo@gmail.com} \\\And
  Geewook Kim$^\ast$$^\dagger$\\
  NAVER Cloud AI\\
  KAIST AI\\
  \texttt{gwkim.rsrch@gmail.com} \\
  }

\begin{document}
\maketitle

\begingroup
\renewcommand\thefootnote{}
\footnotetext{$^{\ast}$ Sukmin Seo and Geewook Kim contributed equally to this work and share first authorship.}
\footnotetext{$^{\dagger}$ Corresponding author.}
\endgroup

\begin{abstract}
Temporal grounding---returning the interval $[t_s, t_e]$ for a
natural-language query over a video---is the language interface to
long-form video, yet has been studied on short videos;
the dynamics of hour-scale natural-language grounding remain
underexplored. We take the position that at hour-scale,
\textbf{the binding constraint is search, not recognition}:
Video-LLMs are bottlenecked not by \emph{localizing} a nearby
event, but---given a natural-language query---by \emph{searching}
for the relevant region of a long video. To test this, we release
\textbf{ExtremeWhenBench}, the first open hour-scale grounding
benchmark (2{,}273 queries over 194 videos, mean 75.7\,min,
max 9\,hr) with an open-form query distribution. Every open Video-LLM collapses while a frame-level retrieval
baseline \emph{outperforms} them; a failure taxonomy attributes
\textbf{85\% of failures to search}; and a retrieve-then-ground
hybrid recovers 6.7$\times$ over the monolithic Video-LLM---mirroring
retrieve-then-read in open-domain QA.
\end{abstract}

\section{Introduction}
\label{sec:intro}

\begin{figure}[t]
\centering
\includegraphics[width=0.95\linewidth]{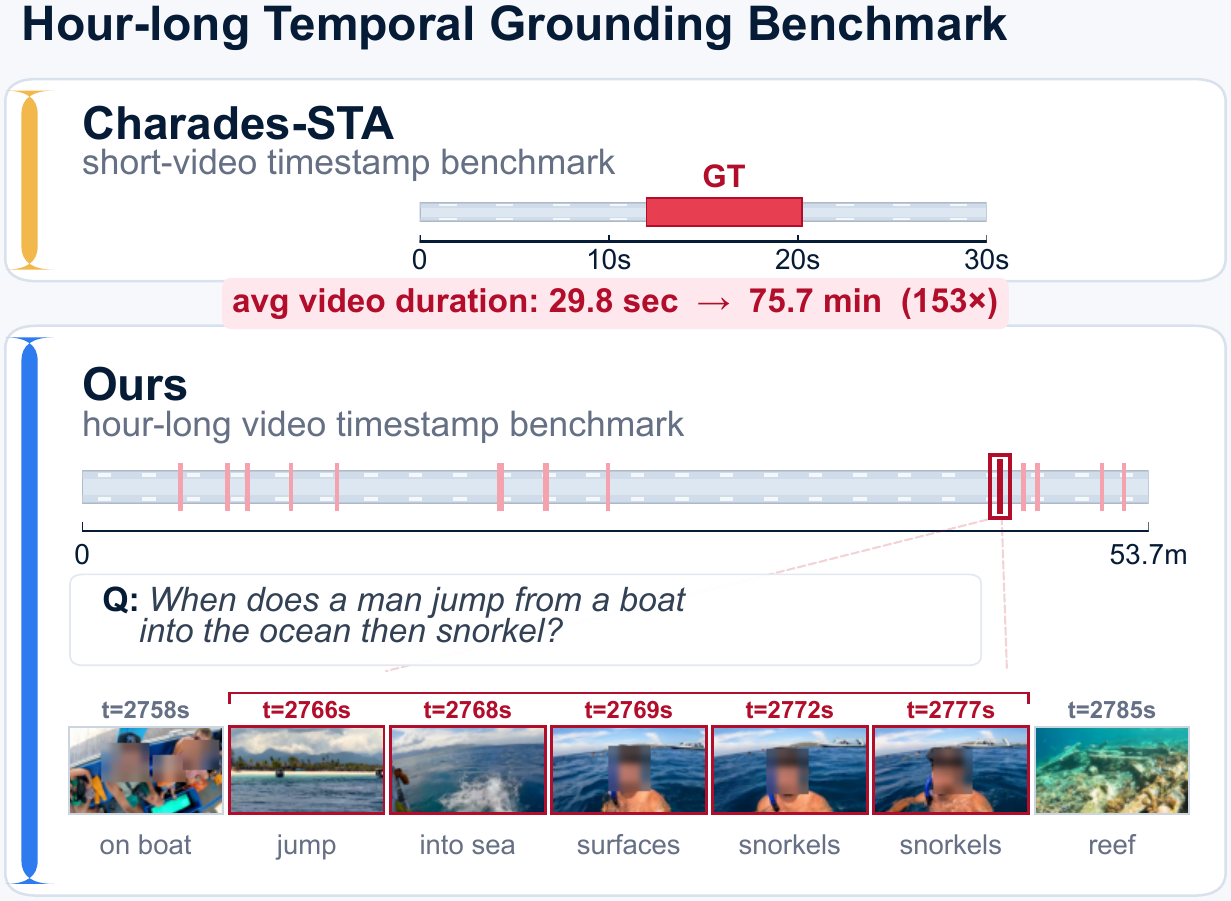}
\caption{ExtremeWhenBench places a $\sim$9\,s event inside a
76\,min video---a search space 153$\times$ larger than Charades-STA
at \emph{matched} event grain. Grounding no longer reduces to
recognition.}
\label{fig:teaser}
\end{figure}

A user asks, in plain language, \emph{``when in this lecture does the
instructor first introduce backpropagation?''}, or \emph{``what
minute of the meeting did we agree on the budget?''}. Returning the
right interval $[t_s, t_e]$ for a natural-language query $q$ over a
video $V$---temporal grounding---is the language interface to
long-form video (Figure~\ref{fig:teaser}). Such queries operate on
\emph{hour-scale} content, yet the grounding literature has been
built on short-video
benchmarks
\citep{gao2017tall,krishna2017densecap,lei2021qvhighlights,regneri2013tacos}
of 30\,s to a few minutes; how Video-LLMs handle hour-scale
natural-language queries remains underexplored.

\paragraph{Why the gap is structural.}
MAD \citep{soldan2022mad} aligns natural-language sentences with
movie audio descriptions but ships only pre-computed CLIP features,
blocking modern Video-LLM evaluation; Ego4D NLQ
\citep{grauman2022ego4d} requires an NDA and a multi-TB download.
Neither is integrated into \texttt{lmms-eval}
\citep{zhang2024lmmseval} or \texttt{VLMEvalKit}
\citep{duan2024vlmevalkit}. 
TVGBench \citep{wang2026timer} is a compact LVLM-friendly benchmark
of 11 balanced query types over short-video sources, preserving the
short-video regime rather than the hour-scale setting we target.
Even recent grounding splits such as TVBench
\citep{cores2025tvbench} remain template-bound, with five 4-gram
prefixes covering 99.9\% of its \texttt{action\_localization} split
(\S\ref{sec:bench})—unlike how users phrase real natural-language
queries.
No existing benchmark
simultaneously satisfies (i) long video length ($\geq$30\,min mean),
(ii) open-form natural-language queries, and (iii) open access via
public URLs, compatible with standard Video-LLM toolkits.

\paragraph{The question hour-scale grounding makes central.}
When a Video-LLM fails on a long video, is it failing to
\emph{recognize} the target event once nearby, or failing to
\emph{find} the region of the long video that the query refers to?
Short-video benchmarks cannot distinguish these because their search
spaces are too small. Hour-scale grounding makes the distinction
central and lets us decompose long-form grounding into a
\emph{search} stage (query-conditioned retrieval over long temporal
context) and a \emph{localize} stage (boundary placement within a
short window). This is structurally the \emph{retrieve-then-read}
decomposition that reshaped open-domain QA
\citep{karpukhin2020dpr,izacard2021leveraging}: the reader's
ceiling is set by the retriever. At hour-scale, the grounding
model's ceiling is set the same way.

\paragraph{Contributions.}
\begin{enumerate}
\item \textbf{ExtremeWhenBench}: an open hour-scale benchmark of
\textbf{natural-language temporal queries} (2{,}273 queries / 194
videos, mean 75.7\,min, max 9\,hr) built on LVBench
\citep{lvbench}, MLVU \citep{mlvu}, and VideoMME \citep{videomme},
with a VLM-in-the-loop boundary verifier that refines 20.6\% of
caption-derived intervals. More details will be available at GitHub\footnote{\url{https://github.com/naver-ai/ExtremeWhenBench}}.
\item A query-side quality filter that produces
\textbf{open-form (not template-bound) natural-language queries},
validated by a $\sim$3$\times$ type-token-ratio (TTR) gap over
Charades-STA and a $\sim$25$\times$ per-question 4-gram-stem gap
over TVBench \citep{cores2025tvbench}; full table in
Appendix~\ref{app:linguistic}.
\item An empirical decomposition of long-form grounding into
search/localize, with a retrieve-then-ground hybrid recovering
6.7$\times$ over the monolithic Video-LLM, mirroring open-domain
QA's retrieve-then-read at hour-scale.
\end{enumerate}

\section{ExtremeWhenBench}
\label{sec:bench}
 
\begin{figure*}[t]
\centering
\includegraphics[width=1.0\linewidth]{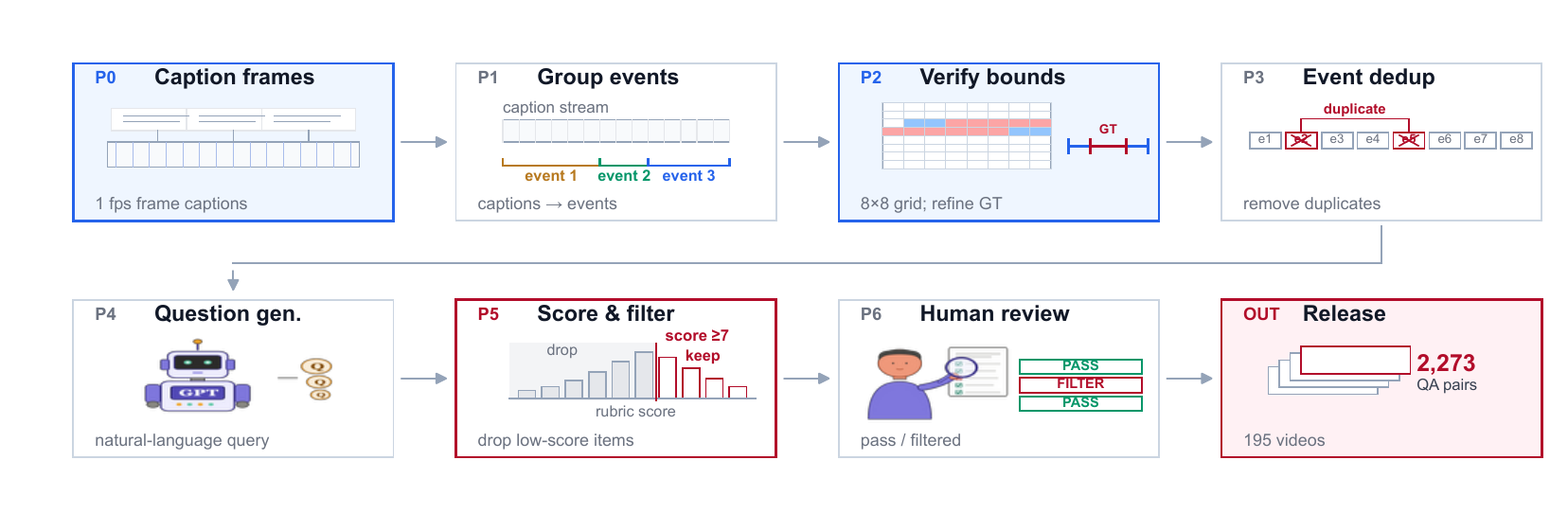}
\vspace{-3.5em}
\caption{Seven-stage benchmark construction. Funnel:
41{,}139 P2-verified events $\rightarrow$ 37{,}599 after P3 dedup
$\rightarrow$ 7{,}188 after P5 pre-pass $\rightarrow$ 2{,}375 after
P5 main filter $\rightarrow$ \textbf{2{,}273} after P6 human review.
Stage details in App.~\ref{app:pipeline}.}
\label{fig:pipeline}
\end{figure*}
 
\paragraph{Why not just pad short videos?}
Padding Charades-STA 30\,s clips with random footage is a simpler
route, but the queries are too broad: Qwen3.5-9B drops from
0.572\,mIoU on native Charades to 0.247 at 10\,min padded and 0.183
at 20\,min (vs.\ 0.369 on a same-length crop of our natural videos),
and 47\%/74.5\% of the worst-200 failures at 10/20\,min place the
prediction \emph{inside} the padding region---an alternative scene
typically matches a query like \emph{``a person walks through the
doorway''} as well as the true Charades segment. We build on naturally
long videos, where each (query, interval) pair is verified unique
within its source video.
 
\paragraph{Source corpora.}
We source 195 videos (mean 75.7\,min, max 9.04\,hr) from LVBench,
MLVU, and VideoMME---all publicly available under their original licenses, with stable URLs,
integrated into \texttt{lmms-eval} / \texttt{VLMEvalKit}. We use
them as \emph{video sources only}; the original QA annotations
are not used.
 
\paragraph{Pipeline.}
The 7-stage pipeline (Figure~\ref{fig:pipeline}; full details in
App.~\ref{app:pipeline}) runs in three phases. \emph{Event mining}
(P0--P3): 1\,fps Qwen3-VL-8B \citep{qwen3vl} captions $\rightarrow$
\texttt{gpt-5-mini} event grouping $\rightarrow$ \texttt{gpt-5.1}
visual boundary verification on $8{\times}8$ frame mosaics
(3.3\% rejected; among kept, 79.4\% verified / 20.6\% refined)
$\rightarrow$ removes repeated events within the same video
(3-aspect match; 41{,}139 $\rightarrow$ 37{,}599). \emph{Question generation} (P4):
\texttt{gpt-5-mini} writes one 8--18-word natural-language
question per event, banning cinematic vocabulary and applying
explicit ordinal disambiguation. \emph{Quality control} (P5--P6):
a \texttt{gpt-5.4} visual-rubric filter scores each question on
validity, uniqueness-within-grids, and temporal difficulty
(anti-CLIP), keeping the top 30\% per group with score $\geq$7
(2{,}375 candidates); CLIP ViT-L/14-336 top-10 coverage then
flags 164 ambiguity-prone items for author review, with 102
dropped to yield the 2{,}273 released items.
 
\paragraph{Statistics.}
The released set is 2{,}273 questions over 194 videos. The median GT
event duration is 9\,s, closely matched to Charades-STA's 7.1\,s, so
the 9\,s event sits inside a 76\,min video---a search space
$\sim$153$\times$ larger at matched event grain.
 
\paragraph{Query-side diversity.}
Long-form grounding is only well-posed if the queries are not
template-bound. We compare against Charades-STA and TVBench
\texttt{action\_localization} (closest video-LLM grounding split)
using sample-size-invariant MATTR \citep{covington2010mattr} and
unique 4-gram \emph{stems} (first four tokens, a
template-diversity probe). Our
queries achieve MATTR 0.78 vs.\ 0.60 (TVBench) and 0.54
(Charades-STA), and yield 1{,}578 unique 4-gram stems---a
$\sim$25$\times$ per-question gap over TVBench, whose five
prefixes cover 99.9\% of its split. We treat this query-side
diversity, not video length, as the benchmark's \emph{primary
linguistic contribution}; long-form video provides the evaluation
substrate (full table in App.~\ref{app:linguistic}).

\section{Empirical Study}
\label{sec:empirical}

\paragraph{Setup.}
We evaluate four open Video-LLMs---Qwen3.5-9B \citep{qwen3},
InternVL3.5-8B \citep{wang2025internvl35}, LLaVA-OneVision-7B
\citep{li2024llavaonevision}, LLaVA-NextVideo-7B
\citep{zhang2024llavanextvideo}---served via vLLM with
\texttt{lmms-eval}; three closed models (GPT-5.4 with $8{\times}8$
grid, Gemini-2.5-flash and Gemini-3.5-flash with raw video); and a
frame-level
language-image retrieval baseline, CLIP ViT-L/14-336
\citep{radford2021clip} at 1\,fps. Frame counts $N$ are swept per
model up to context limit; we report the best-$N$ configuration.
All systems emit a start/end timestamp pair; parse failures count
as IoU=0. We report mIoU here; R@$\tau$, parse-failure rates, and
CLIP smoothing/clipping are in Appendix~\ref{app:metrics}.

\begin{table}[t]
\centering
\small
\begin{tabular}{lrrr}
\toprule
Model & Charades & Ours & Ratio \\
\midrule
Qwen3.5-9B & \textbf{0.579} & \textbf{0.110} & 5.3$\times$ \\
InternVL3.5-8B & 0.359 & 0.003 & 120$\times$ \\
LLaVA-OneVision-7B & 0.226 & 0.003 & 75$\times$ \\
LLaVA-NextVideo-7B & 0.089 & 0.001 & 89$\times$ \\
\midrule
GPT-5.4 (64f) & 0.299 & 0.013 & 23$\times$ \\
Gemini-2.5-flash (1k\,f) & 0.308 & 0.053 & 5.8$\times$ \\
Gemini-3.5-flash (auto-fps) & 0.466 & 0.115 & 4.1$\times$ \\
\midrule
\textbf{CLIP ViT-L/14-336} & 0.332 & \textbf{0.269} & 1.2$\times$ \\
\bottomrule
\end{tabular}
\caption{\textbf{Hour-scale collapse and retrieval cross-over.} Best
per-model mIoU on Charades-STA vs.\ ours for the same
natural-language temporal grounding task. Open Video-LLMs collapse
5--120$\times$; CLIP (frame-level retrieval) \emph{outperforms}
all open Video-LLMs.}
\label{tab:rq12}
\end{table}

\begin{figure}[t]
\centering
\begin{adjustbox}{max width=\linewidth}
\begin{tikzpicture}
\begin{axis}[
  name=mainaxis,
  width=7.2cm, height=3.3cm,
  scale only axis=true,
  xmode=log, log basis x={2},
  xmin=3, xmax=3000,
  xlabel={Frames $N$ (log scale)},
  xtick={4,16,64,256,1024,2048},
  xticklabels={4,16,64,256,1k,2k},
  ymin=0, ymax=0.68,
  ytick={0,0.2,0.4,0.6},
  ylabel={\color{blue!70!black} Charades-STA mIoU},
  axis y line*=left,
  axis x line*=bottom,
  ymajorgrids=true,
  yticklabel style={/pgf/number format/fixed, /pgf/number format/precision=2,
                    font=\footnotesize, blue!70!black},
  xticklabel style={font=\footnotesize},
  ylabel style={font=\footnotesize},
  xlabel style={font=\footnotesize},
  enlarge x limits=0.04,
  legend style={font=\footnotesize,
                at={(0.5,1.04)}, anchor=south,
                draw=none, fill opacity=0.9,
                legend columns=2, column sep=12pt,
                /tikz/every even column/.append style={column sep=12pt},
                inner sep=2pt},
]
\addplot[mark=*, blue!70!black, thick, mark size=2pt] coordinates {
  (4,0.299) (8,0.421) (16,0.514) (32,0.559)
  (64,0.579) (128,0.575) (256,0.557)
};
\addlegendentry{Charades-STA (left)}
\addlegendimage{mark=square*, red!70!black, thick, mark size=2pt}
\addlegendentry{Ours (right)}
\end{axis}
\begin{axis}[
  at=(mainaxis.south west), anchor=south west,
  width=7.2cm, height=3.3cm,
  scale only axis=true,
  xmode=log, log basis x={2},
  xmin=3, xmax=3000,
  xtick=\empty,
  ymin=0, ymax=0.13,
  ytick={0,0.03,0.06,0.09,0.12},
  ylabel={\color{red!70!black} Ours mIoU},
  axis y line*=right,
  axis x line=none,
  yticklabel style={/pgf/number format/fixed, /pgf/number format/precision=2,
                    font=\footnotesize, red!70!black},
  ylabel style={font=\footnotesize},
  enlarge x limits=0.04,
]
\addplot[mark=square*, red!70!black, thick, mark size=2pt] coordinates {
  (8,0.0046) (16,0.0050) (32,0.0066) (64,0.0126)
  (128,0.0215) (256,0.0329) (512,0.0471)
  (1024,0.0676) (2048,0.1095)
};
\end{axis}
\end{tikzpicture}
\end{adjustbox}
\caption{\textbf{Frame-count sweep, Qwen3.5-9B.} Charades-STA mIoU
(blue, left axis) peaks at $N{=}64$ ($0.579$) and stays
flat---more frames carry no new information. Ours mIoU (red, right
axis) rises monotonically and is still climbing at $N{=}2{,}048$
($0.110$); we extend $N$ $8\times$ beyond the Charades regime and
remain unsaturated. Note the $\sim$5$\times$ difference
in y-axis scale.}
\label{fig:sweep}
\end{figure}
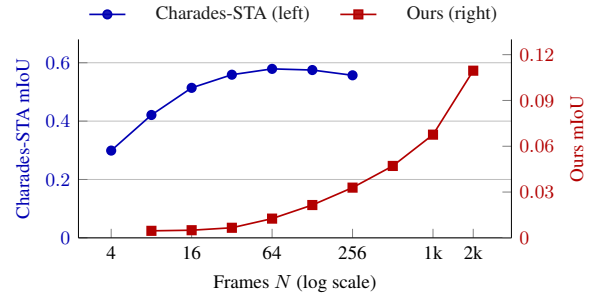

\paragraph{Finding 1: Open Video-LLMs collapse.}
Every open Video-LLM that performs non-trivially on Charades-STA
collapses by at least an order of magnitude on ours
(Table~\ref{tab:rq12}, upper). The smallest relative drop
(5.3$\times$) belongs to Qwen3.5-9B---the only open model whose
context scales gracefully to $N{=}2{,}048$ frames---and its mIoU on
ours rises monotonically from 0.022 at $N{=}128$ to 0.110 at
$N{=}2{,}048$ (Figure~\ref{fig:sweep};
App.~\ref{app:metrics}). \emph{The bottleneck is context
coverage, not recognition}: models that cannot reach the target
frame cannot ground it. Scaling $N$ alone does not close the gap:
even at $N{=}2{,}048$ (0.110), Qwen trails retrieval-alone (0.269).

\paragraph{Finding 2: Query-conditioned retrieval outperforms open
Video-LLMs.}
On Charades-STA the language-image retrieval baseline sits in the
middle of the pack (0.332 vs.\ Qwen 0.579)---recognition matters
most when search is trivial. On ours, the order \emph{flips}: CLIP
(0.269) outperforms the best open Video-LLM (Qwen, 0.110) and every
closed Video-LLM under our compute budget
(Table~\ref{tab:rq12}, last row), pointing to query-conditioned
retrieval over long temporal context as the dominant axis of
difficulty.
\emph{When the task is hour-scale, a model that retrieves the right
region from a natural-language query is more useful than one that
recognizes once nearby.}

\begin{figure}[t]
\centering
\begin{tikzpicture}
\begin{axis}[
  width=\linewidth, height=4.3cm,
  xmode=log, log basis x={10},
  xmin=22, xmax=6500,
  ymin=0, ymax=0.66,
  xlabel={Clip length (GT-centered window, log scale)},
  ylabel={mIoU},
  xtick={30,60,300,600,1200,2400,4560},
  xticklabels={30s,1m,5m,10m,20m,40m,Full},
  xticklabel style={font=\footnotesize},
  yticklabel style={font=\footnotesize,/pgf/number format/fixed,/pgf/number format/precision=2},
  ytick={0,0.2,0.4,0.6},
  xlabel style={font=\footnotesize},
  ylabel style={font=\footnotesize},
  legend style={font=\scriptsize,
                at={(0.03,0.03)}, anchor=south west,
                draw=none, fill opacity=0.92,
                row sep=-1.5pt, inner sep=2pt},
  ymajorgrids=true,
]
\addplot[mark=*, blue!70!black, thick, mark size=1.6pt] coordinates {
  (30,0.572) (60,0.541) (300,0.466)
  (600,0.381) (1200,0.265) (2400,0.098) (4560,0.053)
};
\addlegendentry{Qwen3.5-9B ($N{=}768$)}
\addplot[mark=triangle*, green!55!black, thick, mark size=2pt] coordinates {
  (30,0.578) (60,0.542) (300,0.402)
  (600,0.413) (1200,0.409) (2400,0.320) (4560,0.115)
};
\addlegendentry{Gemini-3.5-flash}
\addplot[mark=square*, red!70!black, thick, mark size=1.6pt] coordinates {
  (30,0.534) (60,0.458) (300,0.387)
  (600,0.358) (1200,0.331) (2400,0.300) (4560,0.269)
};
\addlegendentry{CLIP ViT-L/14-336}
\end{axis}
\end{tikzpicture}
\caption{\textbf{Search--localize cross-over.} mIoU vs.\
GT-centered clip length for Qwen3.5-9B, Gemini-3.5-flash, and a CLIP
frame-level retrieval baseline. Qwen degrades monotonically and
crosses CLIP near a 20\,min clip; Gemini-3.5-flash stays above CLIP
across all windows and only drops below at full-video length.}
\label{fig:crossover}
\end{figure}
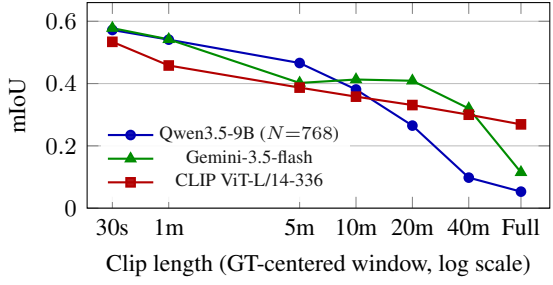

\paragraph{Finding 3: Search--localize cross-over shifts with
Video-LLM strength.}
Cropping the input to a GT-centered window of varying width
(Figure~\ref{fig:crossover}) varies search difficulty without
changing localization difficulty. Both Video-LLMs at a 30\,s
window match their Charades-STA localizer ceiling
(Qwen 0.572, Gemini-3.5 0.578 vs.\ 0.579), so localization is
intact. The Video-LLM--CLIP cross-over then shifts with model
capacity: Qwen crosses CLIP near a 20\,min clip (between $\pm$5
and $\pm$10\,min, with CLIP winning by up to 22 mIoU points at
full video), while Gemini-3.5-flash stays above CLIP across all
shorter windows and only drops below at full-video length
(76\,min). \emph{This locates the regime boundary
mechanistically:} once the clip is long enough that search
matters, recognition stops being the binding constraint---and
the clip length at which this happens scales with the Video-LLM's
long-context capacity.

\begin{table}[t]
\centering
\small
\begin{adjustbox}{max width=\linewidth}
\begin{tabular}{lrr}
\toprule
Pipeline & context & mIoU \\
\midrule
Video-LLM Grounding (Qwen3.5-9B) & 76\,min & 0.053 \\
Retrieval (CLIP) + Rule-based Thresholding & 76\,min & 0.269 \\
\textbf{Retrieval + Video-LLM Grounding} & \textbf{6\,min} & \textbf{0.354} \\
\bottomrule
\end{tabular}
\end{adjustbox}
\caption{\textbf{Retrieve-then-ground hybrid} recovers 6.7$\times$
over the monolithic Video-LLM (Qwen3.5-9B at
$N{=}768$, its native frame budget; 0.053 on full 76\,min) and
1.32$\times$ over retrieval-alone, using only 6\,min of
Video-LLM context. Hybrid uses CLIP top-3 $\pm$1\,min, NMS 60\,s;
per-$K$ ablation in Appendix~\ref{app:metrics}
(Table~\ref{tab:rq4-appendix}).}
\label{tab:rq4}
\end{table}

\paragraph{Finding 4: A retrieve-then-ground hybrid recovers most of
the gap.}
A minimal two-stage pipeline---CLIP top-$K$ retrieval from the
natural-language query, then Qwen3.5-9B grounding within the union
of candidate windows (NMS at 60\,s)---achieves \textbf{0.354\,mIoU
with only 6\,min of Video-LLM context} (Table~\ref{tab:rq4}):
\textbf{6.7$\times$ over the monolithic Video-LLM} and 1.32$\times$
over retrieval-alone. Performance peaks at $K{=}3$ and degrades at
$K{=}10$, where the Video-LLM is again forced to search $\sim$20\,min
of context (Appendix~\ref{app:metrics},
Table~\ref{tab:rq4-appendix}).

\section{Analysis}
\label{sec:analysis}

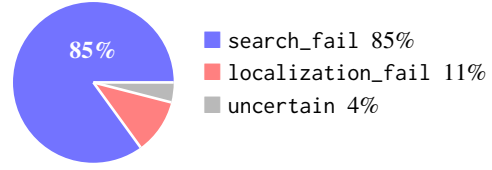
\begin{figure}[t]
\centering
\begin{tikzpicture}[scale=0.85]
\def\R{1.25}
\fill[blue!55] (0,0) -- (\R,0) arc (0:306:\R) -- cycle;
\fill[red!50] (0,0) -- (306:\R) arc (306:345.6:\R) -- cycle;
\fill[gray!55] (0,0) -- (345.6:\R) arc (345.6:360:\R) -- cycle;
\draw[white, line width=1pt] (0,0) -- (0:\R);
\draw[white, line width=1pt] (0,0) -- (306:\R);
\draw[white, line width=1pt] (0,0) -- (345.6:\R);
\node[white, font=\small\bfseries] at (0, 0.50) {85\%};
\node[anchor=west, font=\footnotesize] at (\R+0.3, 0.65)
  {{\color{blue!55}$\blacksquare$}~\texttt{search\_fail}~~85\%};
\node[anchor=west, font=\footnotesize] at (\R+0.3, 0.15)
  {{\color{red!50}$\blacksquare$}~\texttt{localization\_fail}~~11\%};
\node[anchor=west, font=\footnotesize] at (\R+0.3, -0.35)
  {{\color{gray!55}$\blacksquare$}~\texttt{uncertain}~~4\%};
\end{tikzpicture}
\caption{Failure taxonomy on 100 random IoU$<$0.05 cases from
Qwen3.5-9B $N{=}2{,}048$ (1{,}817/2{,}273 fail this threshold;
\texttt{parsing\_fail} and \texttt{refusal} each 0\%).}
\label{fig:failures}
\end{figure}

\paragraph{Failure taxonomy: 85\% search.}
We sample 100 IoU$<$0.05 cases (seed~42) from Qwen3.5-9B
$N{=}2{,}048$ and classify each with a GPT-5.4 classifier into five
a-priori categories (Figure~\ref{fig:failures};
\texttt{parsing\_fail} and \texttt{refusal} are 0\%); we split
\texttt{search\_fail} vs.\ \texttt{localization\_fail} at a
$\pm$5\,min prediction-to-GT offset. \emph{Search failures
(85\%) dominate localization failures (11\%)}: when the model
lands in the right region it places boundaries with reasonable
fidelity, but the dominant failure mode is \emph{picking the wrong
window of the long video}---the mechanistic correlate of the
cross-over.

\paragraph{Synthesis.}
Findings~1--4 and the failure taxonomy converge on a single picture.
At hour-scale, the dominant failure of monolithic Video-LLMs on
natural-language temporal queries is \emph{wrong-window
selection}---a search failure, not a recognition failure. This is
what makes a language-image retrieval baseline---no grounding
training, no language model in the loop---competitive on absolute
mIoU, and what makes a two-stage retrieve-then-ground pipeline beat
monolithic grounding under matched compute. Splitting retrieval and grounding into separate stages---rather
than asking one model to do both---mirrors the retrieve-then-read
consensus in open-domain QA.

\section{Conclusion}
\label{sec:conclusion}

We introduced a new hour-scale benchmark
for \emph{natural-language temporal grounding}, and used it to
decompose long-form grounding into \emph{search} and \emph{localize}.
Open Video-LLMs collapse while a frame-level retrieval baseline
matches or exceeds them; a search--localize cross-over exists at
clip lengths that scale with Video-LLM capacity; 85\% of failures
are search failures; and a minimal
retrieve-then-ground hybrid recovers 6.7$\times$ over the monolithic
Video-LLM with strictly less Video-LLM compute. The natural next
step is to replace the retrieval stage with a stronger
temporally-aware, language-grounded retriever and ask whether the
hybrid's mIoU ceiling rises in tandem with retriever quality. Our
benchmark provides the testbed.

\section*{Limitations}
\label{sec:limitations}

Our pipeline relies on a single source-VLM (Qwen3-VL-8B-Instruct)
for the 1\,fps caption stream; a cross-VLM caption sanity check
that would verify the caption distribution is not biased toward
one model family's failure modes is deferred. The released
benchmark also inherits the genre distribution of LVBench, MLVU,
and VideoMME---long movies, documentaries, and multi-domain
long-form videos---so performance on egocentric or surveillance
content, which may interact differently with the search vs.\
recognition decomposition, is not separately measured. For
query-side diversity, the two long-form references that match our
length regime (MAD and Ego4D NLQ) are NDA/license-gated, so direct
per-query MATTR or distinct-$n$ comparisons against them remain
inaccessible and we rely on short-form references (TVBench,
Charades-STA) for the diversity claim.

On the empirical side, the failure taxonomy in \S\ref{sec:analysis}
uses a single $\pm$5\,min boundary between search and
localization failures; sensitivity to $\pm$2 / $\pm$10\,min is
deferred to follow-up analysis and we expect the qualitative
85/11 split to be robust but the exact share to shift. Finding~4
also uses a single retriever (CLIP ViT-L/14-336): a stronger
temporally-aware, language-grounded retriever could in principle
shift the hybrid's operating point and the location of the
cross-over in Figure~\ref{fig:crossover}, and the relationship
between retriever quality and hybrid ceiling is the natural
follow-up our benchmark is designed to support.

\bibliography{acl_latex_main}

\appendix

\section{Related Work and Positioning}
\label{app:related}

This appendix expands the brief related-work paragraph in
\S\ref{sec:intro} along three lines: (A.1) Video-LLMs trained or
adapted for temporal grounding, (A.2) long-form video benchmarks,
and (A.3) two-stage retrieve-then-X pipelines.

\subsection{Video-LLMs for temporal grounding}
\label{app:related:vllm}

A recent survey \citep{wu2025vtgsurvey} catalogues over 30
Video-LLMs trained or adapted for temporal grounding. We summarize
representative systems in Table~\ref{tab:vllm-taxonomy}. The
mechanisms differ substantially---multi-stage training with text
timestamps \citep{huang2024vtimellm}, timestamp-aware encoders
\citep{ren2024timechat}, continuous-time tokens
\citep{qian2024momentor}, causal event modeling
\citep{guo2025trace}, fully text-to-text grounding
\citep{wang2024hawkeye}---but the reported evaluation splits
cluster on short- to medium-form sources (Charades-STA $\sim$30\,s,
ActivityNet Captions $\sim$2\,min, QVHighlights $\sim$150\,s,
YouCook2 cooking clips). We are not aware of any of these systems
reporting hour-scale natural-language grounding numbers under
standard Video-LLM evaluation toolkits
\citep{zhang2024lmmseval,duan2024vlmevalkit}. Our
\S\ref{sec:empirical} suggests this evaluation choice obscures
the regime where these models break: at hour-scale, the search
bottleneck becomes binding.

\begin{table}[t!]
\centering
\small
\begin{adjustbox}{max width=\linewidth}
\begin{tabular}{lll}
\toprule
System & Venue & Grounding mechanism \\
\midrule
VTimeLLM \citep{huang2024vtimellm} & CVPR24 & Boundary-aware 3-stage training \\
TimeChat \citep{ren2024timechat} & CVPR24 & Timestamp-aware encoder + sliding Q-Former \\
Momentor \citep{qian2024momentor} & ICML24 & Continuous-time tokens (TPM) \\
TRACE \citep{guo2025trace} & ICLR25 & Causal event modeling, task-interleaved heads \\
HawkEye \citep{wang2024hawkeye} & arXiv24 & Fully text-to-text, coarse-grained spans \\
\bottomrule
\end{tabular}
\end{adjustbox}
\caption{Representative Video-LLMs trained or adapted for temporal
grounding. All reported eval splits are short- to medium-form
(Charades-STA, ActivityNet Captions, QVHighlights, YouCook2);
none extends to our 75.7\,min mean regime.}
\label{tab:vllm-taxonomy}
\end{table}

\subsection{Long-form video benchmarks}
\label{app:related:longform}

Several prior benchmarks address aspects of long-form temporal
grounding but each fails at least one of our three constraints
(long mean length, open-form natural-language queries, open
access via public URLs compatible with standard Video-LLM
toolkits):

\noindent\textbf{MAD} \citep{soldan2022mad}: hour-scale movie-derived
queries, but ships only pre-computed CLIP features rather than raw
video, precluding modern Video-LLM evaluation.

\noindent\textbf{Ego4D NLQ} \citep{grauman2022ego4d}: hour-scale
egocentric video with NL queries, but access requires an NDA and a
multi-TB download.

We further use LVBench \citep{lvbench}, MLVU \citep{mlvu}, and
VideoMME \citep{videomme} as raw \emph{video sources} only; their
original QA annotations (multiple-choice or generative) do not
enter our pipeline. 
To our knowledge, our benchmark is the first
to simultaneously satisfy $\geq$30\,min mean length, open-form
natural-language queries, and open access via public URLs
compatible with standard Video-LLM toolkits, such as, \texttt{lmms-eval}~\citep{zhang2024lmmseval}.

\subsection{Two-stage retrieve-then-X pipelines}
\label{app:related:twostage}

Two-stage retrieve-then-read is the default architecture for
open-domain QA over long text
\citep{karpukhin2020dpr,izacard2021leveraging,lewis2020rag},
with the consistent finding that end-to-end performance is
bounded by retrieval quality---the structural parallel we test
for long video. The closest video-side counterpart is RGNet
\citep{hannan2024rgnet}, which proposes a unified
clip-retrieval-and-grounding architecture for 20--120\,min
videos, evaluated on Ego4D NLQ and MAD. RGNet \emph{proposes the
architecture}; we \emph{empirically establish} that the
decomposition is needed at all for hour-scale grounding under
standard Video-LLM toolkits (Findings 1, 3, 4 in
\S\ref{sec:empirical}), using a minimal training-free hybrid
(CLIP retrieval + a frozen Video-LLM grounder) to make the
point. The two contributions are complementary.

\section{Annotation pipeline details}
\label{app:pipeline}

\begin{figure*}[!htbp]
\centering
\includegraphics[width=\textwidth]{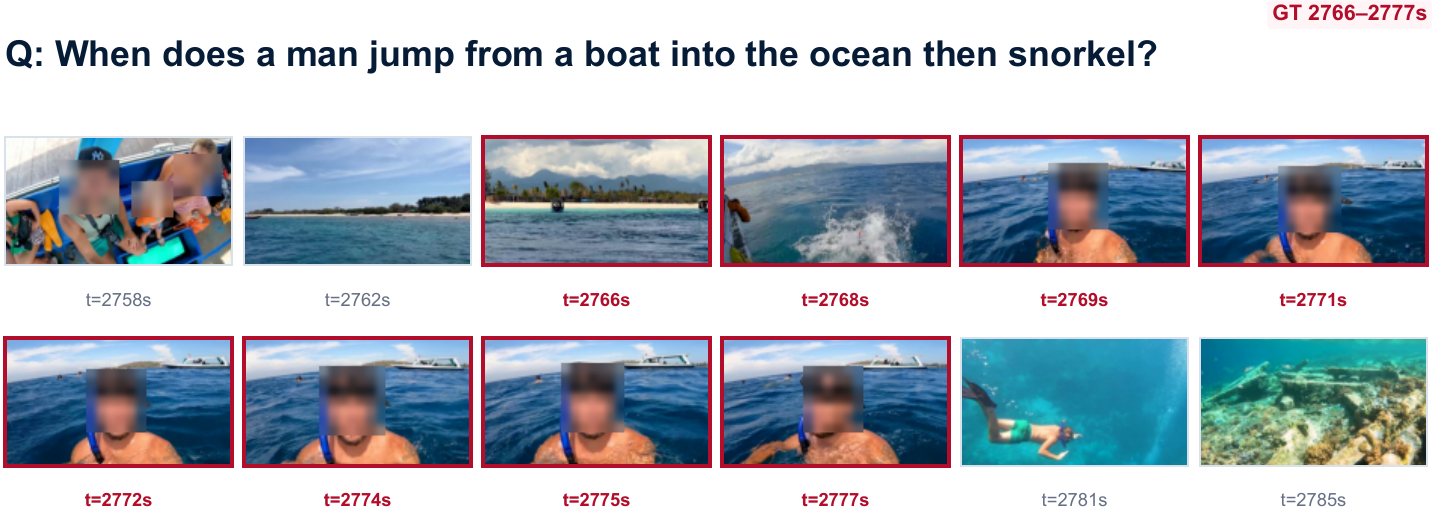}\\[6pt]
  \caption{A qualitative example.}
\label{fig:ex-bench-gallery-1}
\end{figure*}

We document the seven stages summarized in \S\ref{sec:bench} and
Figure~\ref{fig:pipeline} in per-stage form below.
Table~\ref{tab:pipeline-detail} lists the model, reasoning setting,
and prompt used at each stage; ``reasoning'' refers to the
OpenAI-style reasoning-effort parameter, which we set to
\texttt{medium} throughout. P0 is the only stage that consumes raw
video frames at scale; downstream stages operate on the caption
stream emitted by P0 except for P2 (boundary verification), which
re-consults the original video frames in an $8{\times}8$ grid with
$\pm$5\,s padding around the candidate event.
Figure~\ref{fig:ex-bench-gallery-1} shows the natural-language query, the
ground-truth interval, and a strip of frames sampled within and
around the GT.

\begin{table}[t!]
\centering
\small
\begin{adjustbox}{max width=\linewidth}
\begin{tabular}{llll}
\toprule
Stage & Model & Reasoning & Prompt \\
\midrule
P0 caption (1\,fps) & Qwen3-VL-8B & --- & frame-level desc \\
P1 event grouping & gpt-5-mini & medium & V1d \\
P2 boundary verify & gpt-5.1 & medium & v3 ($8{\times}8$+pad) \\
P3 within-vid dedup & gpt-5-mini & medium & v2 \\
P4 question gen & gpt-5-mini & medium & v2 (8--18 tok) \\
P5 quality filter & gpt-5.4 & medium & 3-criteria, top-30\% \\
P6 human review & authors & --- & CLIP-flagged 164 \\
\bottomrule
\end{tabular}
\end{adjustbox}
\caption{Per-stage configuration.}
\label{tab:pipeline-detail}
\end{table}

\paragraph{P0 Captioning.}
1\,fps frame-level descriptions are generated by Qwen3-VL-8B-Instruct
\citep{qwen3vl}. Each frame yields a short text describing visible
entities and actions, with the absolute video timestamp attached.
The caption stream is the input substrate for P1--P5.

\paragraph{P1 Event grouping.}
\texttt{gpt-5-mini} reads the caption stream of each video and emits
a sequence of candidate events, each with a coarse $[t_s, t_e]$
boundary derived from adjacent captions that describe the same
ongoing action. This step is purely textual and does not re-consult
the video.

\paragraph{P2 Boundary verification.}
\texttt{gpt-5.1} re-examines each candidate by sampling 64 frames
uniformly from a $\pm$5\,s-padded window around the candidate
interval and tiling them into a single $8{\times}8$ mosaic image
(64 thumbnails $=$ one ``frame grid'' the VLM can see at once). For
each grid the VLM returns one of four verdicts: \textsc{Verified}
(boundaries are correct as proposed), \textsc{Refined} (boundaries
shifted to better match the event onset/offset), \textsc{Split}
(the candidate contains two distinct events and is divided), or
\textsc{Rejected} (the candidate does not depict the proposed
event). Among kept events the verdict mix is 79.4\%
\textsc{Verified} / 20.6\% \textsc{Refined}; a further $\sim$3.3\%
of raw P1 candidates are \textsc{Rejected}, and \textsc{Split} is
rare ($\sim$0\%).

\paragraph{P3 Within-video deduplication.}
One \texttt{gpt-5-mini} call per video reads all P2-verified events
and flags a pair as duplicate \emph{only when at least three
semantic aspects of the event description overlap} (e.g., same
actor \emph{and} same action \emph{and} same setting). Pairs that
share only one aspect---for instance, the same character appearing
in two different scenes, or two events in the same setting with
different actions---are kept as distinct. The pass retains 91.4\%
of P2 events (41{,}139 $\rightarrow$ 37{,}599), confirming that
within long videos true near-duplicates are rare.

\paragraph{P4 Question generation.}
\texttt{gpt-5-mini} writes one natural-language question per
deduplicated event under three constraints: (i) length 8--18 words;
(ii) ban on cinematic vocabulary (e.g., ``scene'', ``cut'',
``establishing shot'') that leaks production-side framing into a
viewer-side query; (iii) explicit ordinal disambiguation
(e.g., ``the first time'', ``when finally'') to prevent referent
ambiguity for queries that admit multiple candidate intervals
within the same video.

\paragraph{P5 Visual-rubric quality filter.}
Two passes feed the released number. The \emph{pre-pass}
(\texttt{gpt-5-mini}, text-only) drops obviously bad items
(unanswerable, non-memorable, ambiguous, borderline-duplicate,
over-long) from the P4 output and applies a 25\,s GT-duration
cutoff, leaving 7{,}188 candidates. The \emph{main pass} groups
neighbouring questions per video (8--15 per group; the group ends
when the next GT is more than 300\,s away) and builds 5--15
$8{\times}8$ frame grids that cover the union of those GTs
($\pm$10\,s pad, 1\,fps, empty time skipped; each thumbnail keeps
its absolute video timestamp). \texttt{gpt-5.4} via Batch API
scores each question 1--10 on (A) \textsc{validity}---does the
marked interval really answer the question; (B) \textsc{uniqueness
within grids}---is there a different moment in the shown frames
that would answer it equally well; (C) \textsc{temporal difficulty}
(anti-CLIP)---down-weight questions a single still frame can solve,
up-weight those that need motion or sequence. We keep the top 30\%
per group, then drop any remaining item with score~$<7$, producing
2{,}375 release candidates. Kept-score histogram:
7$:$209 / 8$:$1{,}394 / 9$:$751 / 10$:$21.

\paragraph{P6 Targeted human review.}
For each P5-passing query we compute CLIP ViT-L/14-336 top-10
candidate windows ($\pm$1\,min, NMS 60\,s) and flag the 164 items
($\approx$6.9\% of the 2{,}375-item pool) whose GT interval is not
covered by any top-10 hit---a strong signal of annotation
ambiguity, missing visual evidence, or wording-induced referent
ambiguity. At the final stage, the authors manually inspect all
164 (video, query, GT) triples and discard 102 (ambiguous boundary,
event missing from frames, or wording artifact), yielding the
2{,}273-item released set spanning 194 videos (one source video
lost all its queries at this stage).

\paragraph{Cost.}
Total API cost was $\approx$\$200 across the 195-video corpus.
P2 (boundary verification on
the $8{\times}8$ grid) accounts for the largest fraction; P0
captioning is a single-VLM cost amortized across all downstream
stages.

\paragraph{License.}
Source videos and original annotations retain their upstream
licenses (LVBench, MLVU, VideoMME); our added annotations
(event boundaries, queries, quality scores) will be released
under MIT.

\section{Full per-model metrics}
\label{app:metrics}

\begin{table}[t!]
\centering
\small
\begin{adjustbox}{max width=\linewidth}
\begin{tabular}{lrrrrrr}
\toprule
$N$ & 4 & 16 & 64 & 128 & 256 & best \\
\midrule
\multicolumn{7}{l}{\textit{Charades-STA mIoU}} \\
Qwen3.5-9B & .299 & .514 & \textbf{.579} & .575 & .557 & $N{=}64$ \\
InternVL3.5 & .298 & \textbf{.359} & .335 & .336 & .004$^\dagger$ & $N{=}16$ \\
LLaVA-OV-7B & .234 & .206 & OOM & --- & --- & $N{=}8$ \\
LLaVA-NextV & \textbf{.089} & .075 & .072 & 0 & --- & $N{=}4$ \\
\midrule
$N$ & 128 & 256 & 512 & 1024 & 2048 & best \\
\midrule
\multicolumn{7}{l}{\textit{Ours mIoU}} \\
Qwen3.5-9B & .022 & .033 & .047 & .068 & \textbf{.110} & $N{=}2048$ \\
InternVL3.5 & .003 & 0 & --- & --- & --- & sat. \\
LLaVA-OV-7B & 0 & --- & --- & --- & --- & sat. \\
\bottomrule
\end{tabular}
\end{adjustbox}
\caption{Per-model frame-count sweep on Charades-STA (top) and ours
(bottom). $^\dagger$context overflow.}
\label{tab:metrics-sweep}
\end{table}

\paragraph{Retrieve-then-ground: per-$K$ ablation.}
Table~\ref{tab:rq4-appendix} expands the hybrid row of
Table~\ref{tab:rq4} across retrieval breadth. Recall is the fraction
of items whose GT interval is covered by at least one of the top-$K$
candidate windows; the VLM (Qwen3.5-9B $N{=}768$) is invoked only on
the union of those windows. Performance peaks at $K{=}3$: $K{=}10$
adds recall but forces the VLM to again search $\sim$20\,min, while
$K{=}1$ has too low coverage.

\begin{table}[t!]
\centering
\small
\begin{adjustbox}{max width=\linewidth}
\begin{tabular}{lrrr}
\toprule
Hybrid setting & VLM ctx & Recall & mIoU \\
\midrule
CLIP top-1 $\pm$2.5\,min & 5\,m & 62.8 & 0.302 \\
\textbf{CLIP top-3 $\pm$1\,min (NMS 60\,s)} & \textbf{6\,m} & \textbf{81.6} & \textbf{0.354} \\
CLIP top-10 $\pm$1\,min & 20\,m & 96.7 & 0.295 \\
\bottomrule
\end{tabular}
\end{adjustbox}
\caption{Per-$K$ ablation of the retrieve-then-ground hybrid.
Recall is the fraction of items whose GT is covered by the top-$K$
candidate set after NMS.}
\label{tab:rq4-appendix}
\end{table}

\paragraph{Open models: where the sweep saturates.}
Two patterns are visible in Table~\ref{tab:metrics-sweep}. On
Charades-STA, every model peaks at small $N$ ($N{=}4$ to $N{=}64$) and
degrades thereafter, consistent with the 30\,s clip length: more
frames do not add information. On ours, only Qwen3.5-9B makes use of
larger $N$; InternVL3.5 and LLaVA-OneVision saturate near $N{=}128$
because their context windows do not support deeper sweeps without
overflow. LLaVA-NextVideo is omitted from the Ours sweep because its
Charades-STA peak (0.089 at $N{=}4$) is already too weak.

\paragraph{Closed models and CLIP baseline.}
GPT-5.4 ($8{\times}8$ grid, 64f) scores 0.013\,mIoU on ours;
expanding to 512f across eight grids improves to 0.042\,mIoU at
50.7\% parse-failure rate. Gemini-2.5-flash (1024f via raw video
file) scores 0.053\,mIoU; Gemini-3.5-flash with auto-fps
on the same input reaches \textbf{0.115\,mIoU}, the best closed-model
result---still below CLIP-alone (0.269). CLIP ViT-L/14-336 (1\,fps,
peak $\beta{=}0.7$, clip$[4,60]$) scores 0.269\,mIoU with no parse
failures; the absence
of prompting and response parsing contributes to robustness at
hour-scale.

\paragraph{Window-clipping sweep (numerical).}
Table~\ref{tab:rq3-appendix} reports the per-window mIoU values
visualized in Figure~\ref{fig:crossover}. Qwen3.5-9B runs at
$N{=}768$. Gemini-3.5-flash uses fps=5 up to $\pm$5\,min and auto-fps
(VideoMetadata fps=0; Gemini samples internally) for $\pm$10\,min,
$\pm$20\,min, and Full.

\begin{table}[t!]
\centering
\small
\begin{adjustbox}{max width=\linewidth}
\begin{tabular}{llrrr}
\toprule
Window & Clip & Qwen & Gemini-3.5 & CLIP \\
\midrule
$\pm$15\,s    & 30\,s   & 0.572 & \textbf{0.578} & 0.534 \\
$\pm$0.5\,min & 1\,min  & 0.541 & \textbf{0.542} & 0.458 \\
$\pm$2.5\,min & 5\,min  & \textbf{0.466} & 0.402 & 0.387 \\
$\pm$5\,min   & 10\,min & 0.381 & \textbf{0.413} & 0.358 \\
$\pm$10\,min  & 20\,min & 0.265 & \textbf{0.409} & 0.331 \\
$\pm$20\,min  & 40\,min & 0.098 & \textbf{0.320} & 0.300 \\
Full          & 76\,min & 0.053 & 0.115 & \textbf{0.269} \\
\bottomrule
\end{tabular}
\end{adjustbox}
\caption{Window-clipping sweep, human-filtered ($n{=}2{,}273$).
\textbf{Bold} = best per row.}
\label{tab:rq3-appendix}
\end{table}

\section{Linguistic diversity: full table}
\label{app:linguistic}

\paragraph{Setup.}
We compare against two reference points: Charades-STA test
(canonical moment-retrieval benchmark) and TVBench
\texttt{action\_localization} \citep{cores2025tvbench} (closest
video-LLM grounding split). We report TTR; MATTR
\citep{covington2010mattr} (sliding-window TTR, sample-size-invariant
at $W{=}50$); Distinct-$n$ \citep{li2016diversity}; and unique 4-gram
\emph{stems} (the first 4 tokens of each question), which directly
probe question-template diversity.

\begin{table}[t!]
\centering
\small
\begin{adjustbox}{max width=\linewidth}
\begin{tabular}{lrrr}
\toprule
Metric & Ours & TVBench & CSTA \\
\midrule
TTR & 0.173 & 0.091 & 0.058 \\
MATTR (W{=}50) & \textbf{0.780} & 0.595 & 0.539 \\
Distinct-1 & 0.173 & 0.091 & 0.058 \\
Distinct-2 & \textbf{0.610} & 0.230 & 0.249 \\
Distinct-3 & \textbf{0.845} & 0.331 & 0.506 \\
Distinct-4 & \textbf{0.946} & 0.392 & 0.705 \\
Unique bigrams (tot) & 17{,}618 & 556 & 1{,}690 \\
Unique 4-gram stems & \textbf{1{,}578} & \textbf{5} & 707 \\
Top-5 stem coverage & 47\% & 99.9\% & 19\% \\
\bottomrule
\end{tabular}
\end{adjustbox}
\caption{Full diversity table. ``Top-5 stem coverage'' is the
fraction of questions whose 4-gram prefix matches one of the five
most common prefixes. TVBench's 99.9\% reflects its declared
5-template design.}
\label{tab:linguistic-full}
\end{table}

\paragraph{Findings.}
On sample-size-invariant MATTR, our queries (0.78) exceed both
TVBench (0.60) and Charades-STA (0.54). The most striking gap is at
the 4-gram level: \emph{TVBench's five 4-gram prefixes
(\texttt{during which part}, \texttt{when in the}, \texttt{at what
moment}, \texttt{in the given}, \texttt{can you identify}) cover
99.9\% of its 160-question split}, while our top-5 prefixes cover
only 47\% of 2{,}273 questions, yielding 1{,}578 unique stems
($\sim$25$\times$ the per-question rate). The two long-form
references that match our length regime (MAD, Ego4D NLQ) are
NDA/license-gated, so we cite only their published statistics
(MAD \citep{soldan2022mad}: $>$60K unique words across 384K
queries).

\end{document}